# Predicting wind pressures around circular cylinders using machine learning techniques

Gang Hu and K.C.S. Kwok

*Centre for Wind, Waves and Water, School of Civil Engineering, The University of Sydney, Sydney, NSW 2006, Australia*
*gang.hu@sydney.edu.au; kenny.kwok@sydney.edu.au*

January 20, 2019

**Abstract**

Numerous studies have been carried out to measure wind pressures around circular cylinders since the early 20[th] century due to its engineering significance. Consequently, a large amount of wind pressure data sets have accumulated, which presents an excellent opportunity for using machine learning (ML) techniques to train models to predict wind pressures around circular cylinders. Wind pressures around smooth circular cylinders are a function of mainly the Reynolds number (*Re*), turbulence intensity (*Ti*) of the incident wind, and circumferential angle of the cylinder. Considering these three parameters as the inputs, this study trained two ML models to predict mean and fluctuating pressures respectively. Three machine learning algorithms including decision tree regressor, random forest, and gradient boosting regression trees (GBRT) were tested. The GBRT models exhibited the best performance for predicting both mean and fluctuating pressures, and they are capable of making accurate predictions for *Re* ranging from $10^4$ to $10^6$ and *Ti* ranging from 0% to 15%. It is believed that the GBRT models provide very efficient and economical alternative to traditional wind tunnel tests and computational fluid dynamic simulations for determining wind pressures around smooth circular cylinders within the studied *Re* and *Ti* range.

## 1. Introduction

Circular cylinders can be classified as a type of bodies in between streamlined bodies such as airfoils, and non-streamlined bodies with sharp edges such as rectangular prisms. Subjected to an oncoming flow, streamlined bodies avoid flow separation, while non-streamlined bodies exhibit flow separations at fixed points (e.g. the sharp corners). For circular cylinders, the flow separation is much more complicated. The separation position varies according to a number of parameters, including incident flow velocity, turbulence, body geometry, and surface roughness (Niemann and Hölscher, 1990). Consequently, the flow around a circular cylinder is one of the most challenging problems in fluid dynamics. However, understanding the flow regime and aerodynamic characteristics of circular cylinders is crucial for ensuring the safety of engineering structures ranging from heat exchangers to bridge cables, silos, industrial chimney, and cooling towers, as shown in Fig. 1.

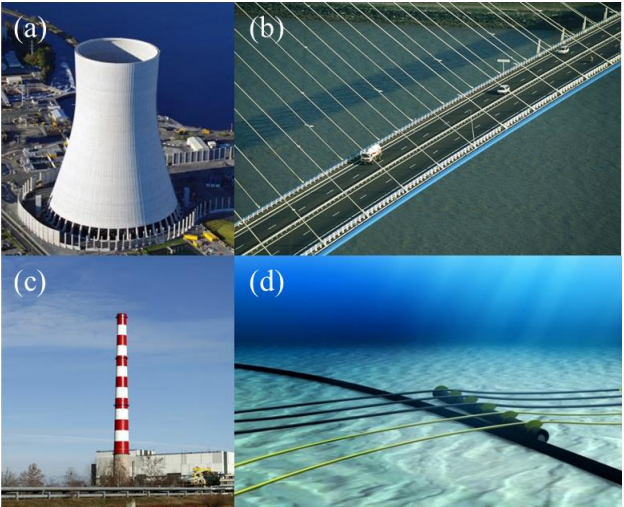

Fig. 1 Engineering structures with circular cross sections: (a) cooling tower, (b) cables of bridge, (c) industrial chimney, (d) underwater cables.

Since the early 20$^{th}$ century, a considerable amount of research has been devoted to study flow around circular cylinders (e.g. Thom, 1933; Zdravkovich, 1997a, 1997b). The research methodologies adopted by these studies can be categorized into wind tunnel experiments (Ke et al., 2017; Zou et al., 2018), field measurements (Zhao et al., 2017, 2016), and computational fluid dynamic simulations (Catalano et al., 2003; Moussaed et al., 2014; Yeon et al., 2016; Zhao and Cheng, 2011). Although substantial efforts have been made to acquire the pressure distributions around circular cylinders with various dimensions under various incident flow conditions, time-consuming and expensive wind tunnel tests and/or computational fluid dynamic simulations remain indispensable for determining the pressure around a circular cylinder with a particular dimension immersed in a particular flow field.

Fortunately, the research endeavors in the past have accumulated a large amount of wind pressure data sets for circular cylinders. These data sets form a basis for using machine learning (ML) techniques to build models for predicting pressures around circular cylinders. ML is a powerful tool of data mining that automates analytical model building. It is a branch of artificial intelligence based on the idea that systems can learn from data and make decisions or predictions with minimal human intervention. There are a number of ML techniques such as supervised ML (Kotsiantis, 2007), semi-supervised ML, and unsupervised ML (Hastie et al., 2009a). Supervised ML relies on an algorithm to generate a hypothesis or model to predict future values based on input values. Supervised ML requires labelling of data. This means that data would need to be separated into two categories called labels and features. For unsupervised ML, none of the data is labeled. The supervised ML algorithms mainly include k-Nearest Neighbors, decision trees, naïve Bayes, logistic regression, support vector machines, and neural network (Harrington, 2012; Hastie et al., 2009b). The general process of applying supervised ML to a problem is given by Kotsiantis (2007) as shown in Fig. 2.

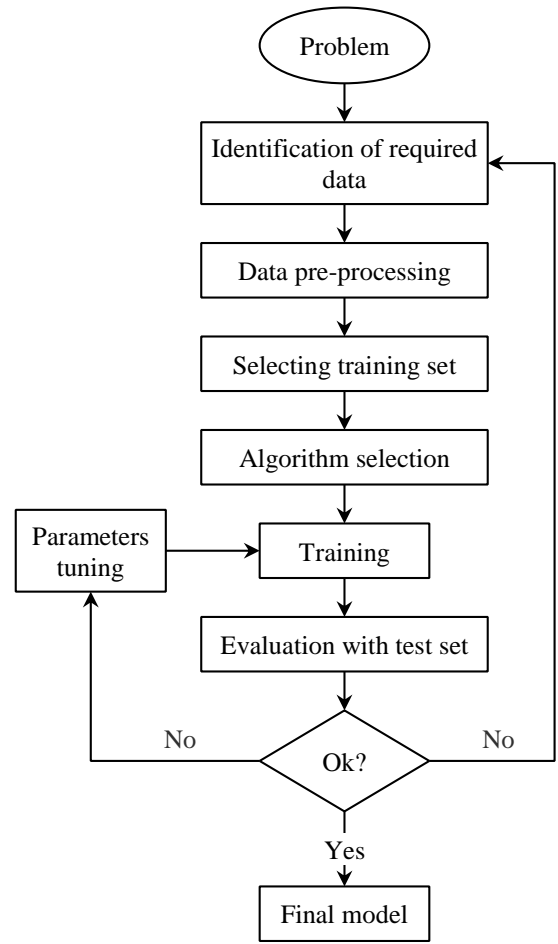

Fig. 2 General process of applying supervised machine learning to a problem (After Kotsiantis, 2007).

ML techniques have shown a huge potential of application in various engineering fields, such as structural health monitoring (Figueiredo et al., 2011; Ni et al., 2005; Nick et al., 2009; Santos et al., 2016; Worden and Manson, 2007), construction materials (Cheng et al., 2012; Chou et al., 2014; Sonebi et al., 2016), wind energy (Becker and Thrän, 2017; Clifton et al., 2013; Heinermann and Kramer, 2016), and transportation engineering (Liu et al., 2018a, 2018b). However, the application of ML techniques in wind engineering is still in its infancy. Wu and Kareem (2011) utilized an artificial neural network framework with the embedded cellular automata scheme to develop a new promising approach to model aerodynamic nonlinearities in the time domain. Aruljayachandran et al. (2016) assessed the full-scale acceleration data obtained from field measurements at six levels of the world's tallest building, Burj Khalifa, by using ML techniques. Li et al. (2018) adopted a data-driven approach using a ML scheme to model vortex-induced vibrations (VIVs) of a suspension bridge based on a database of field measured VIVs of the bridge over six years. A decision tree algorithm was adopted to train the VIV mode classification model and a support vector regression (SVR) algorithm was used to model the VIV response of the bridge deck. The classification and regression models can accurately identify and predict the VIV response for various modes of the bridge. Another interesting study done by Jin et al. (2018) built a data-driven model for predicting the velocity field around a circular cylinder via fusion convolutional neural networks based on measured pressure field on the cylinder. The model was proven to be accurate when compared with CFD results and furthermore it successfully learned the underlying flow regimes of the cylinder.

This study aims to build ML regression models for predicting mean and fluctuating pressures around smooth circular cylinders under various combinations of Reynolds number and turbulence intensity in the range with sufficient training data sets. Section 2 provides a literature review on aerodynamic characteristics of circular cylinders and introduces data collected from previous studies. In section 3, ML algorithms adopted in this study are detailed and the prediction models based on these algorithms are built. Performances of the ML models are compared in section 4. Additionally, predictions on mean and fluctuating pressures by using the best models are demonstrated in this section. Discussions and conclusions are given in sections 5 and 6 respectively.

## 2. Literature review and data collection

As mentioned above, flow around circular cylinders is very complex due to the variable separation point, and hence the surface pressure is complex. Overall, the surface pressure around a smooth cylinder is a function of mainly Reynolds number, turbulence intensity of incident wind, and circumferential angle of circular cylinder.

### 2.1. Effects of Reynolds number

Reynolds number ($Re$) is a dimensionless parameter defining the ratio of fluid inertia force to viscous forces. It is an important parameter to describe the flow patterns around circular cylinders and also dominate aerodynamic characteristics and surface pressures of the cylinders. The effects of $Re$ on the force coefficients acting on circular cylinders are summarized and depicted in Fig. 3. Evidently, the effects of $Re$ on these coefficients are very complex.

Due to the importance of $Re$ in the flow around a circular cylinder, it attracts many studies. In 1930s, Thom (1933) adopted an arithmetical method to solve the equations of viscous flow past a circular cylinder at $Re$ =20, and obtained its drag and pressure distribution. Meanwhile, the surface pressure of the cylinder at very low $Re$ ranging from 3.5 to 240 were measured. To understand the aerodynamic behaviors of a circular cylinder at very high $Re$ from $10^6$ to $10^7$, Roshko (1961) measured the pressure distribution, force coefficients, and Strouhal numbers of a large circular cylinder in a pressurized wind tunnel. Another set of experiments at high $Re$ flow around a circular cylinder was also conducted in a pressurized wind tunnel by Achenbach (1968). The drag coefficient, pressure distribution, and separation point position were determined in the $Re$ range from $6\times10^4$ to $5\times10^6$ in that study. Jones et al. (1969) measured the force coefficients acting on and surface pressures around a circular cylinder in a two-dimensional flow at $Re$ from $3.6\times10^5$ to $1.87\times10^7$ in a transonic wind tunnel. For the critical $Re$ flow around a circular cylinder, Farell and Blessmann (1983) experimentally studied the pressure distribution around and the velocity fluctuation in the wake of the cylinder. Two sub-states were identified in the critical or lower transition: the first state features symmetric pressure distributions while the second one characterizes by intense flow oscillations. A review of the flow past circular cylinder at different $Re$ was presented by Niemann and Hölscher (1990).

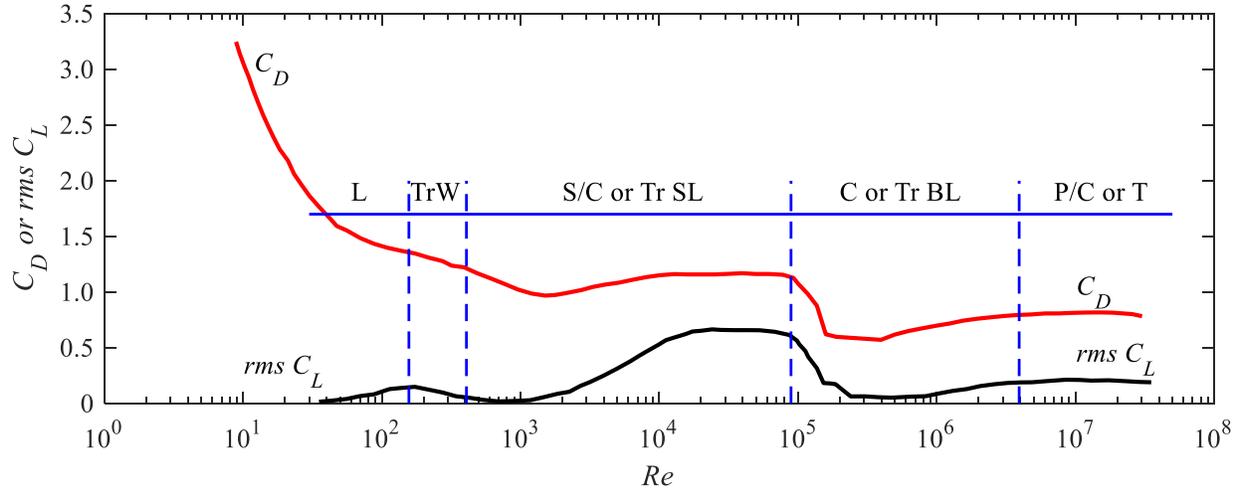

Fig. 3 Effects of Reynolds number on force coefficients acting on circular cylinders; $C_D$ - mean drag coefficient, rms $C_L$ - fluctuating lift coefficient; L – laminar in all regions of flow, TrW – transition in wake, laminar elsewhere, TrSL – transition in free shear layers, wake turbulent, TrBL – transition in boundary layers, T – turbulent in all regions of flow (After Zdravkovich, 1990).

Apart from the wind tunnel experimental approach, field measurement on realistic structures with a circular cross section is another effective way to study high $Re$ flow past circular cylinder. Ruscheweyh (1975) measured the wind pressures around a tapered reinforced-concrete television tower at $Re$ as large as $1.8\times10^7$. Melbourne et al. (1983) made measurements of the wind-induced responses of and surface pressures around a 265 m high reinforced concrete stack at $Re = 2\times10^7$. Waldeck (1989) measured mean pressure distribution around a 300 m concrete chimney. A series of field measurements on wind pressures around cooling towers were conducted by Zhao et al. (2017).

### 2.2. Effects of turbulence intensity of incident wind

Turbulence intensity ($Ti$) is the ratio of fluctuating component to the associated mean component of wind velocity, which quantifies fluctuation of wind flow. As reported by Norberg (1987), the flow around a circular cylinder is insensitive to $Ti$ at $Re$ lower than about 1000, while the influence of $Ti$ is significant at $Re$ higher than about 1000. Surry (1972) experimentally examined the effects of high $Ti$ on the flow past a circular cylinder at the subcritical $Re$ range. Batham (1973) reported that at $Re = 1.11\times10^5$ and $2.35\times10^5$, the presence of turbulence plays a role on suppressing the coherent vortex shedding and leads to a complex pressure field independent of $Re$. For a similar $Re$ range, from $0.8\times10^5$ to $6\times10^5$, Bruun and Davies (1975) found that the pressure fluctuation magnitudes and correlations on the cylinder windward face are strongly associated with free-stream turbulence. However, on the leeward side, the pressure fluctuations are almost independent of the turbulence. Another study concentrated on the effect of the turbulence on the flow past a circular cylinder at subcritical $Re$ ranging from $5.2 \times 10^4$ to $2.09 \times 10^5$ was conducted by Sadeh and Saharon (1982). They concluded that the effect of turbulence in this $Re$ range is equivalent to an increase in the $Re$. That is to say, the existence of incident turbulence alters the surface pressure distribution, shifts the separation point position, and reduces mean drag coefficient. Cheung and Melbourne (1983) studied the effect of turbulence at $Re$ up to $10^6$. Their experimental results showed that the turbulence effects in the subcritical and supercritical $Re$ ranges are opposite. For example, in the subcritical $Re$ range, the base pressure increases while the minimum pressure, mean and fluctuating drag and fluctuating lift decrease with turbulence. However, the effects are opposite in the supercritical $Re$ range. Furthermore, they found that the turbulence effect not only is equivalent to an increase in the $Re$, but also enhances interactions between the separated shear layers and the wake.

## 2.3. Effects of circumferential angle of circular cylinder

Flow past a circular cylinder experiences a complicated evolution process along the circumference of the cylinder, which causes a complex pressure distribution around the cylinder surface. When flow approaches the cylinder, it first comes to rest at the front of the cylinder, i.e., the stagnation point with the maximum pressure as shown in Fig. 4. Further downstream, the flow accelerates, and the surface pressure drops and reaches the minima at a certain circumferential angle $\theta_m$, as shown in Fig. 5. After passing this location with the minimum pressure, the flow slows down and the pressure increases. The point with the local maximum pressure corresponds to the separation point $\theta_s$. The region downstream the separation point is immersed in the wake region that has a nearly constant pressure. Overall, as depicted in Fig. 5, the mean pressure coefficient (mean $Cp$) on the cylinder surface decreases from the maxima at the stagnation point, reaches the minima at $\theta_m$, then increases to the local maxima at $\theta_s$, and keeps nearly invariant in the wake region. Thus, the effect of the circumferential angle on the mean pressure is significant. Likewise, its effect on the fluctuating pressure coefficient (RMS $Cp$) cannot be ignored, as can be seen in Fig. 5.

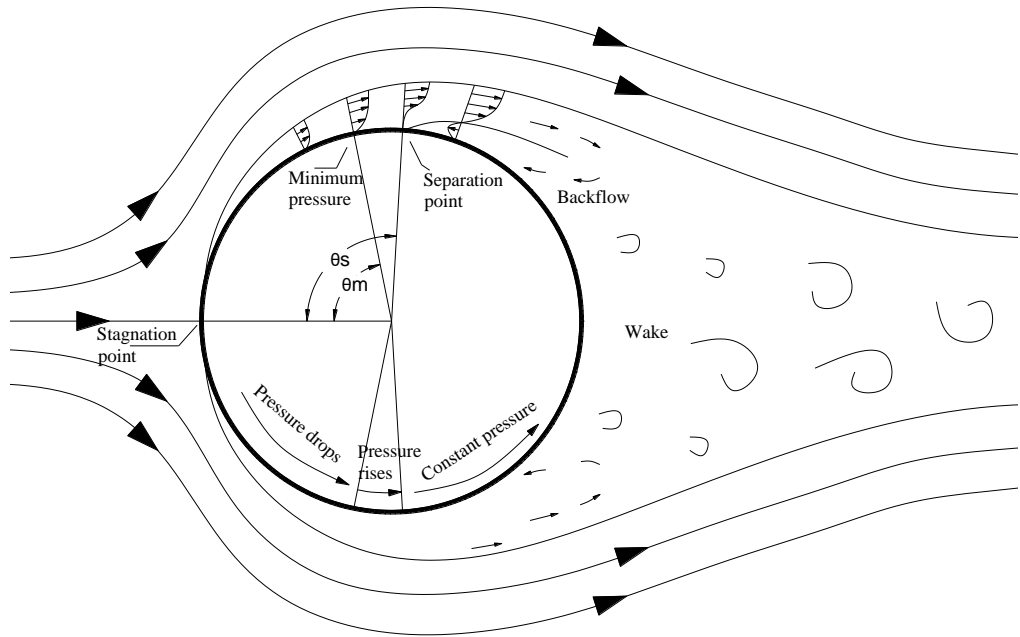

Fig. 4 Detailed flow pattern around a circular cylinder (from unknown source).

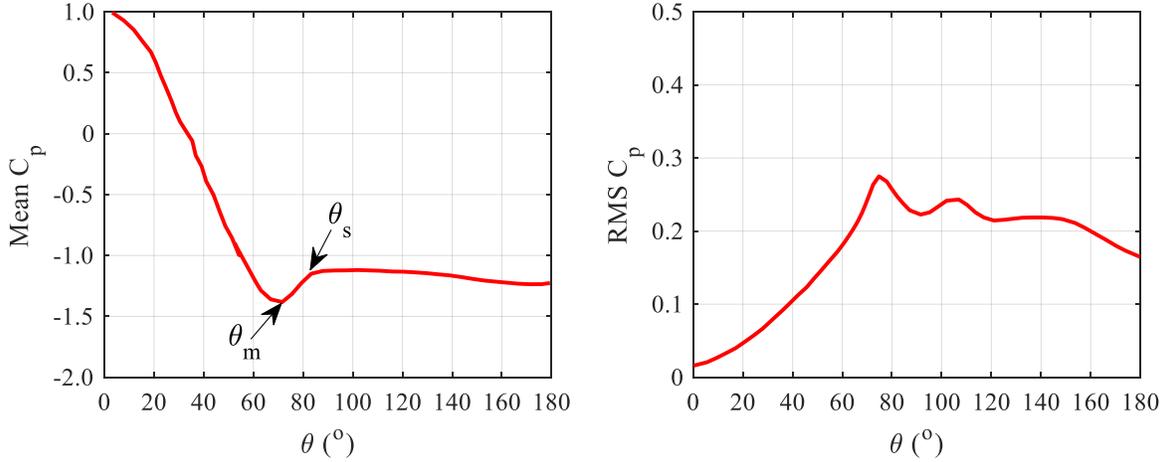

Fig. 5 Variations of mean and fluctuating pressure coefficients with circumferential angle. $\theta_m$ represents the minimum pressure angular position; $\theta_s$ denotes the separation angle. Mean $Cp$: Roshko (1961); fluctuating $Cp$: Batham (1973).

## 2.4. Data collection of surface pressures around circular cylinders

Data collected from the published high-quality experimental studies were used as data set in training ML models for predicting the mean and fluctuating pressures around circular cylinders. The relevant information of these studies used in the present study is summarized in Table 1.

Table 1 Test details of the studies used in the present study

| No. | Experimental studies | Reynolds number | Turbulence intensity (%) | Mean pressure | Fluctuating pressure |
|---|---|---|---|---|---|
| 1 | Lockwood and McKinney (1960) | $5.02\times10^5 \sim 1.064\times10^6$ | 0 | √ | × |
| 2 | Roshko (1961) | $1.1\times10^5 \sim 8.4\times10^6$ | 0 | √ | × |
| 3 | Tani (1964) | $1.06\times10^5 \sim 4.65\times10^5$ | 0 | √ | × |
| 4 | Achenbach (1968) | $1.0\times10^5 \sim 3.6\times10^6$ | 0.7 | √ | × |
| 5 | Jones et al. (1969) | $5.2\times10^5 \sim 1.78\times10^7$ | 0.2 | √ | × |
| 6 | Surry (1972) | $3.38\times10^4 \sim 4.42\times10^4$ | 2.5~14.7 | √ | √ |
| 7 | Batham (1973) | $1.11\times10^5 \sim 2.39\times10^5$ | 0.5; 12.9 | √ | √ |
| 8 | Bruun and Davies (1975) | $8.0\times10^4 \sim 4.8\times10^5$ | 0.2~11.0 | × | √ |
| 9 | Güven et al. (1980) | $4.1\times10^5$ | 0.2 | √ | × |
| 10 | So and Savkar (1981) | $2.62\times10^4 \sim 8.21\times10^5$ | 0; 10 | √ | × |
| 11 | Sadeh and Saharon (1982) | $5.2\times10^4 \sim 2.14\times10^5$ | 0 | √ | × |
| 12 | Kiya et al. (1982) | $2.64\times10^4 \sim 3.98\times10^4$ | 1.4 ~ 12.8 | √ | √ |
| 13 | Arie et al. (1983) | $1.57\times10^5$ | 0.3 | √ | √ |
| 14 | Cheung and Melbourne (1983) | $1.0\times10^5 \sim 4.8\times10^6$ | 0.4 ~ 9.1 | √ | √ |
| 15 | Cheung (1983) | $7.0\times10^4 \sim 1\times10^6$ | 0.4~9.1 | √ | √ |
| 16 | Farell and Blessmann (1983) | $1.27\times10^5 ; 2.32\times10^5$ | 0.4 | √ | × |
| 17 | Melbourne et al. (1983) | $2.4\times10^5 \sim 1.4\times10^7$ | 0.1~17.0 | √ | √ |
| 18 | Cantwell and Coles (1983) | $1.4\times10^5$ | 0.6 | √ | × |
| 19 | Basu (1985) | $2.64\times10^4 \sim 7.1\times10^5$ | 1.4~14.7 | √ | √ |
| 20 | Batham (1985) | $9.89\times10^4 \sim 4.18\times10^5$ | 8.0; 14.0 | √ | × |
| 21 | Norberg (1986) | $4.1\times10^4 \sim 5.5\times10^4$ | 0.1~4.1 | √ | √ |
| 22 | Kwok (1986) | $3.54\times10^4 \sim 2.35\times10^5$ | 0 ~ 12.9 | √ | × |
| 23 | Igarashi (1986) | $6.1\times10^4$ | 0.5 | √ | × |
| 24 | Norberg (1987) | $2.99\times10^3 \sim 8.02\times10^3$ | 0.1; 1.4 | √ | × |

| | | | | | | |
|---|---|---|---|---|---|---|
| 25 | Norberg and Sunden (1987) | $1.33 \times 10^4 \sim 2.30 \times 10^5$ | 0.1; 1.4 | √ | √ | |
| 26 | Higuchi et al. (1989) | $1.90 \times 10^5$ | 0.3 | √ | × | |
| 27 | Yokuda and Ramaprian (1990) | $1.3 \times 10^4 \sim 1.0 \times 10^5$ | 0.7 | √ | √ | |
| 28 | Norberg (1992) | $8.0 \times 10^3 \sim 2.1 \times 10^5$ | 0.1 | √ | √ | |
| 29 | Sun et al. (1992) | $3.25 \times 10^5; 6.25 \times 10^5$ | 0.1 | √ | √ | |
| 30 | Gu et al. (1993) | $1.95 \times 10^5; 6.5 \times 10^5$ | 0.1; 10 | √ | √ | |
| 31 | West and Apelt (1993) | $2.2 \times 10^4 \sim 2.15 \times 10^5$ | 0.2~7.5 | × | √ | |
| 32 | Fox and West (1993) | $4.4 \times 10^4 \sim 6.6 \times 10^4$ | 0.2 | × | √ | |
| 33 | Sakamoto and Haniu (1994) | $6.5 \times 10^4 \sim 1.0 \times 10^5$ | 0.2 | √ | × | |
| 34 | Cheung and Melbourne (1995) | $1.0 \times 10^5; 1.0 \times 10^6$ | 0.4; 9.1 | √ | √ | |
| 35 | Gu (1996) | $2.2 \times 10^5 \sim 4.5 \times 10^5$ | 0.2 | √ | × | |
| 36 | Gu and Sun (1999) | $3.3 \times 10^5$ | 0.2 | √ | × | |
| 37 | Arunachalam et al. (2001) | $1.95 \times 10^5$ | 18.0 | √ | √ | |
| 38 | Gatto et al. (2001) | $6.8 \times 10^4; 9.6 \times 10^4$ | 0.2 | √ | √ | |
| 39 | Nishimura and Taniike (2001) | $6.1 \times 10^4; 1.1 \times 10^5$ | 0.1 | √ | √ | |
| 40 | Eaddy (2004) | $2.10 \times 10^5 \sim 1.12 \times 10^6$ | 4.4; 7.2; 12.4 | × | √ | |
| 41 | Wornom et al. (2011) | $3.9 \times 10^3 \sim 2.0 \times 10^4$ | 0.1 | √ | × | |
| 42 | Gu et al. (2012) | $3.0 \times 10^4 \sim 6.0 \times 10^4$ | 0.1 | √ | × | |
| 43 | Qiu et al. (2014) | $1.66 \times 10^5 \sim 8.28 \times 10^5$ | 0.5 | √ | × | |
| 44 | Min (2014) | $5.5 \times 10^4 \sim 8.87 \times 10^5$ | 0.3~11.2 | √ | × | |
| 45 | Mcclure and Yarusevych (2016) | $6.0 \times 10^3 \sim 1.2 \times 10^4$ | 0.1 | √ | √ | |
| 46 | Cheng et al. (2016) | $2.1 \times 10^5 \sim 4.19 \times 10^5$ | 0.5; 4 | √ | √ | |
| 47 | Gao et al. (2017) | $2.67 \times 10^4$ | 0.4 | √ | √ | |

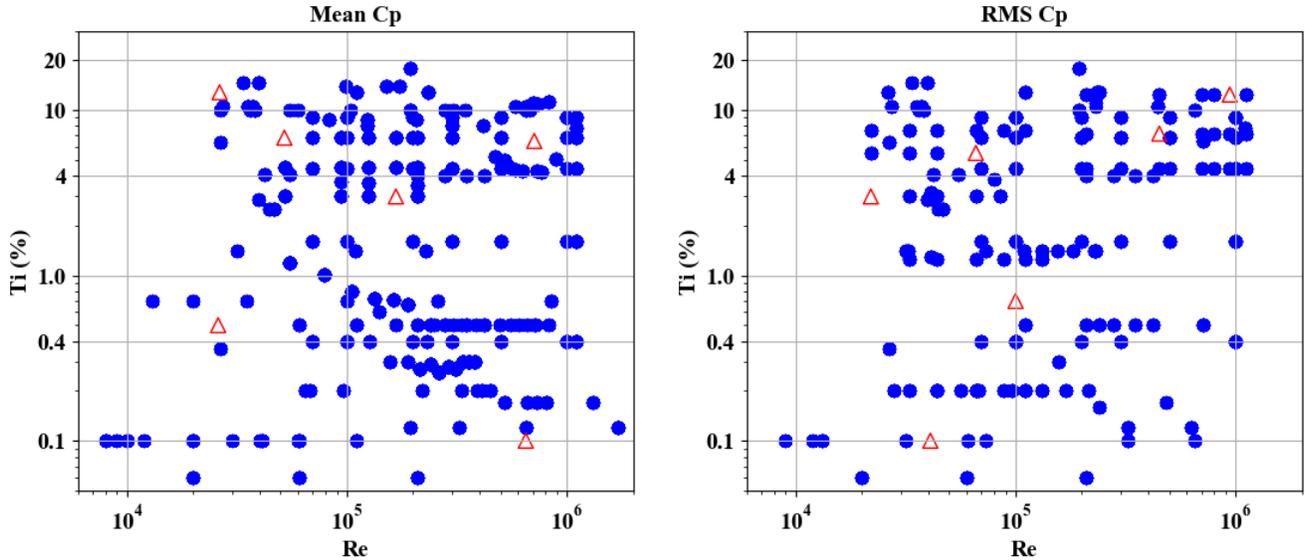

Fig. 6 Data sets of mean pressure coefficient (Mean *Cp*) and fluctuating pressure coefficients (RMS *Cp*) around circular cylinders collected from literature. Red triangular markers: data used to validate the final ML models in section 4.2.

Generally, both the mean and fluctuating pressure at two sides of a cross-section of a circular cylinder is symmetrical with respect to the incident wind direction. In this study, therefore, pressure on only half circle is studied. All studies summarized in Table 1 measured pressures on a half circle or the whole circle of the cylinder section. Therefore, the circumferential angle $\theta$ in these studies covers the range from 0 to 180°. To demonstrate the distribution of the data collected, two scatter plots of the mean pressure and fluctuating pressure data with various combinations of *Re* and *Ti* are given in Fig. 6. The red triangular markers in each plot correspond to the data that will be used to validate the final ML models in section 4.2. They were randomly selected and set aside from the training data set. It can be seen that in the Reynolds number range from $10^4$ to $10^6$ and turbulence intensity range from 0% to 15%, the data sets broadly scatter over the whole range for both mean and fluctuating pressure coefficients. Therefore, it is anticipated that the ML models can cover the above ranges in this study.

## 3. Machine learning algorithms and prediction model training

In this study, three ML algorithms including the decision tree regressor (DTR), random forest (RF), and gradient boosting regression trees (GBRT) were employed to train models for predicting the mean and fluctuating pressure around smooth circular cylinders. The implementation of these algorithms relied on the open-source library *scikit-learn*.

### 3.1. Performance evaluation methods

#### 3.1.1. k-fold cross-validation

There are a number of methods that have been developed to evaluate the performance of ML models, such as re-substitution, hold-out, cross-validation (CV), and bootstrap (Reich and Barai, 1999). Ideally, if there is enough data, it is desirable to set aside a validation data set and use this data set to evaluate the performance of the prediction model, which is hold-out validation (Refaeilzadeh et al., 2009). However, this is often impossible since data are usually scarce. To finesse this problem, the CV method is often used to minimize the bias associated with random sampling of training and hold out data samples. The CV method includes leave-one-out and *k*-fold CV. Most empirical studies found *k*-fold CV method to be reasonably unbiased and with reasonable variability (Reich and Barai, 1999). The *k*-fold CV method splits the data into *k* roughly equal-sized parts (Hastie et al., 2009b). In each of *k* rounds of model training and validation, it chooses a different data subset for testing and trains the model with the remaining *k*-1 data subsets, as shown in Fig. 7. The performance of the trained model is evaluated by the test data subset. The accuracy of the ML algorithm is then expressed as average accuracy of the *k* models in *k* validation rounds. An appropriate value for *k* is crucial for the performance of *k*-fold CV method. Although increasing *k* leads to more performance estimates and large training set size, the overlap between training sets also increases. Considering these competing factors, Refaeilzadeh et al. (2009) suggested that *k* = 10 is a decent compromise. Therefore, 10-fold CV method was adopted to evaluate the performance of the ML algorithms in this study.

#### 3.1.2. 3-stage evaluation process

It is very important to achieve the performance as best as possible, out of the data set when training ML models for practical problems. Therefore, optimizing the hyperparameters of ML algorithms is indispensable during the model training process. Considering the aforementioned *k*-fold CV method and this optimization procedure, a 3-stage evaluation process called training-testing-validation (TTV) was proposed by Reich and Barai (1999). This TTV process comprises of three steps. The first step subdivides the data into data for model training and testing. In this study, 90% of the whole data set were used to train the ML model and the rest were used to test the model. The second step selects the best ML algorithms and tunes hyperparameters using the *k*-fold CV method. The last step builds a model based on the entire training data set by using the best ML algorithms and best hyperparameters. This model is then validated by the testing data set, i.e. 10% of the whole data set. This TTV process and the *k*-fold CV method are summarized in Fig. 7.

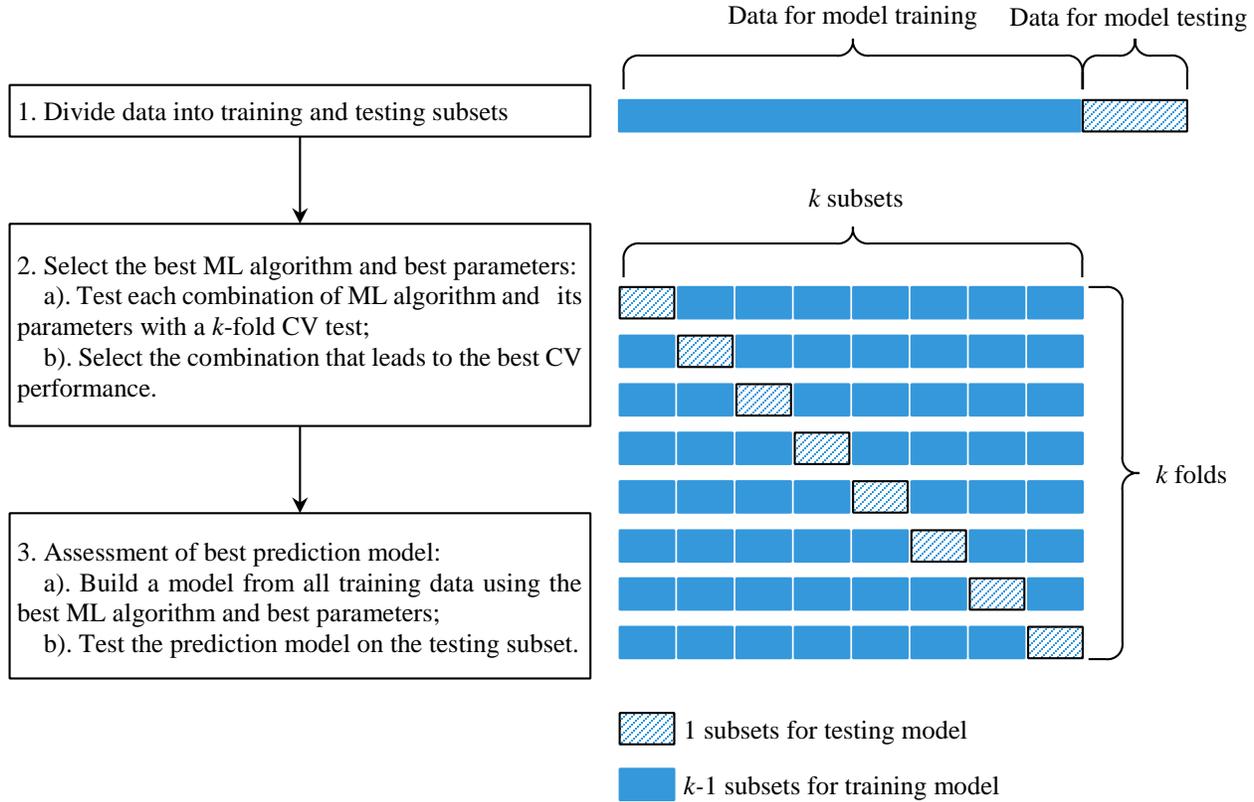

Fig. 7 *k*-fold CV method and 3-stage evaluation process (After Reich and Barai, 1999).

## 3.2. Machine learning algorithms and model trainings

### 3.2.1. Decision tree regression

The decision tree method trains a supervised ML model whereby the local region is identified in a sequence of recursive splits in a smaller number of steps (Alpaydin, 2014). A decision tree consists of root node, internal decision nodes and terminal leaves (see Fig. 9). Given an input, a test is applied at each node, and one of the branches is taken depending on the outcome. This process starts at the root and is repeated recursively until a leaf node is hit, at which point the value written in the leaf constitutes the output.

The classification and regression tree (CART) algorithm, one of the algorithms of implementing decision trees (Breiman, 1984), was utilized to construct decision trees in this study. This algorithm can handle both categorical and numerical dependent variables (Loh, 2014, 2011). That is to say, it is able to construct both classification trees and regression trees depending on the variable type, either categorical or numerical. The Gini index is usually used as its impurity function (Loh, 2011). The Gini index is defined as

$$Gini(t) = 1 - \sum_{i=0}^{c-1}[p(i|t)]^2 \qquad (1)$$

where $p(i/t)$ denotes the fraction of records belonging to class $i$ at a given node $t$; $c$ is the number of classes. During the model training process in the present study, effects of the hyperparameters, including the maximum depth of the tree (max_depth), the maximum leaf nodes, and the minimum number of samples required to be at a leaf node (min_samples_leaf), on the performance of the model for the data set of mean pressure coefficients were evaluated as shown in Fig. 8. It can be seen that when the maximum depth and the maximum leaf nodes exceeds exceed 20 and 1250 respectively, and the min_samples_leaf equals 2, the 10-fold mean squared error (MSE) reaches the minimum. Therefore, 20, 1250, and 2 were chosen for the above three parameters respectively for the decision tress regressor (DTR) for the mean pressure coefficients in this study. The same hyperparameter optimization process was performed on the data set of fluctuating pressure coefficients. It was found 20, 1500, and 2 are the optimal values for the three parameters respectively. An exemplary regression tree for demonstrating the usage of DTR to predict the mean pressure coefficients of circular cylinder is given in Fig. 9.

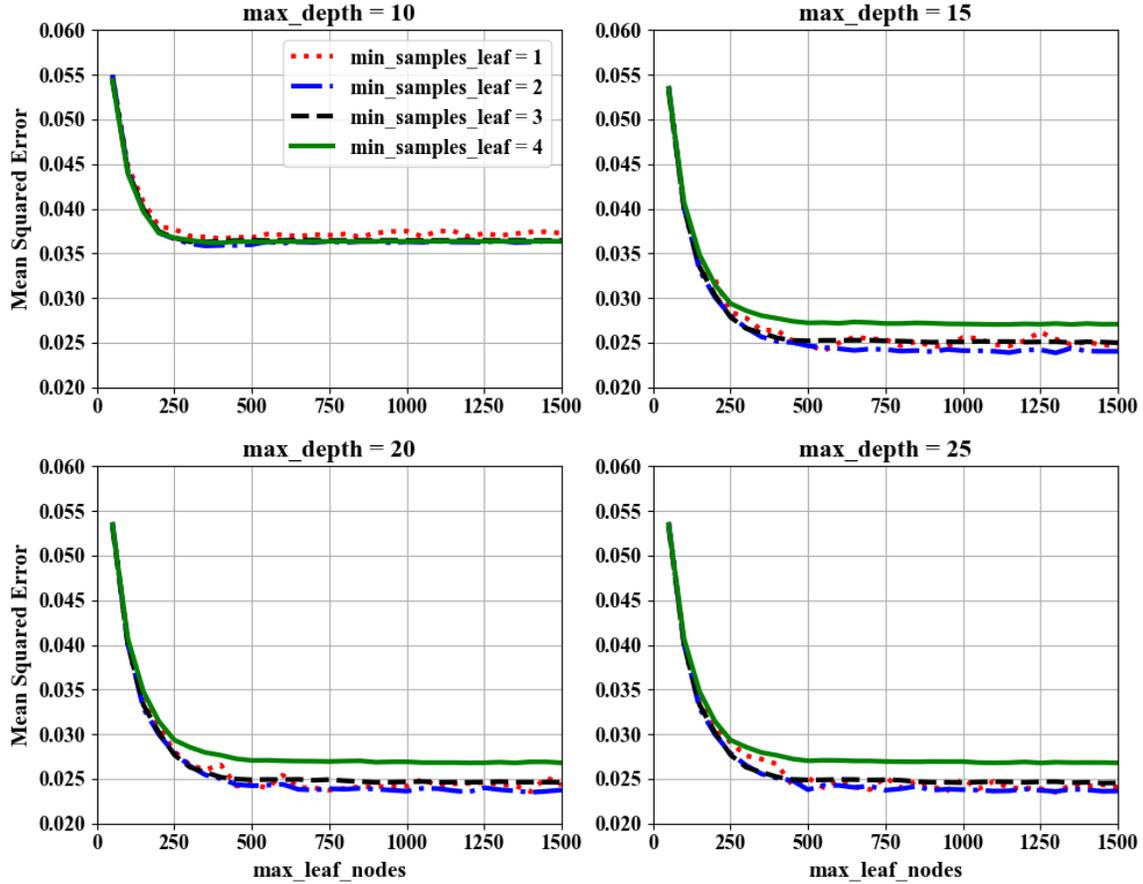

Fig. 8 Variations of 10-fold MSE with hyperparameters of DTR for the data set of mean pressure coefficients.

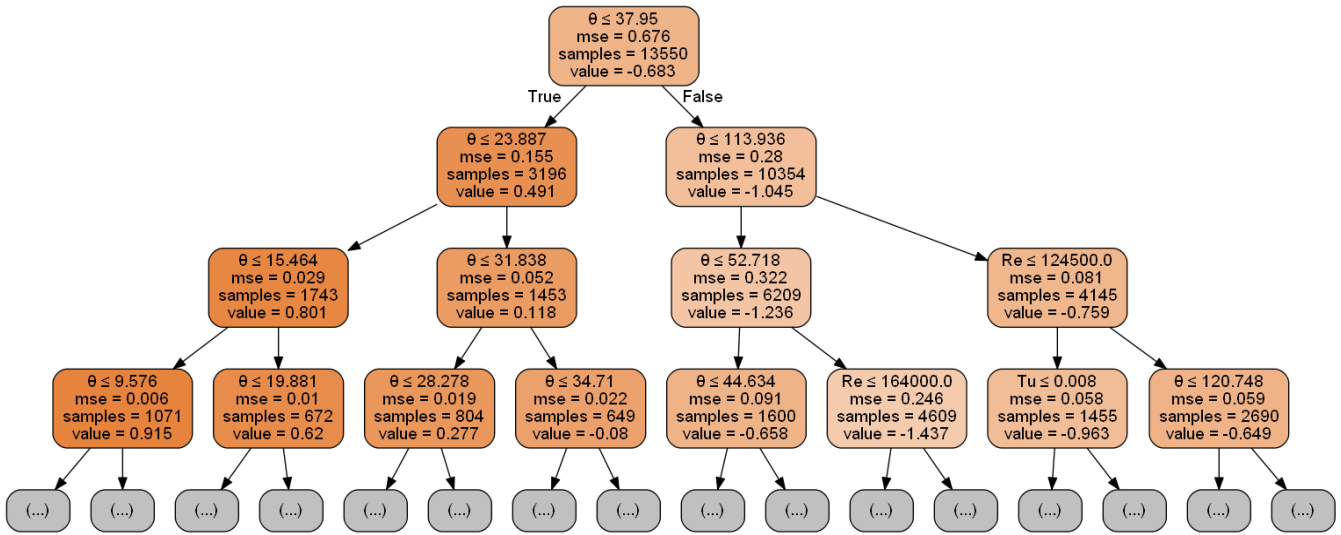

Fig. 9 Exemplary regression tree for demonstrating usage of DTR to predict mean pressure coefficients of circular cylinder. *θ*: circumferential angle; MSE: mean squared error; *Re*: Reynolds number; *Ti*: turbulence intensity.

### 3.2.2. Ensemble methods

It has been recognized that a single learner (e.g., DTR) is sensitive to training data and it is not robust. This can be overcome by aggregating multiple weak learners, which is a so-called ensemble method (Zhou, 2012). The ensemble method constructs a set of weak learners from training data and performs regression by averaging the predictions by each weak learner, as shown in Fig. 10. According to how the weak learners are generated, there are two ensemble methods: parallel ensemble method and sequential ensemble method (Zhou, 2012). In parallel ensemble method with *Bagging* as a representative, the weak learners are generated in parallel, while in sequential ensemble method with *Boosting* as a representative, the weak learners are generated sequentially. It has been proven that an ensemble is usually much more accurate and reliable than a single learner (Lahouar and Ben Hadj Slama, 2017; Ma and Cheng, 2016; Natekin and Knoll, 2013; Wang et al., 2018; Zhou, 2012).

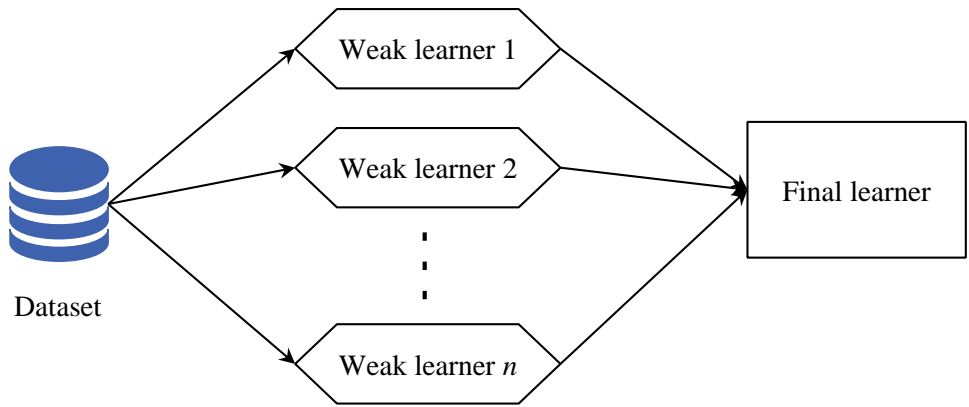

Fig. 10 A common architecture of ensemble method.

*(a) Bagging regression trees: Random Forest (RF)*

Bagging, short for bootstrap aggregating, was introduced by Breiman (1996) and can be applied to regression trees, called bagging regression trees. In the bagging regression trees, a regression tree is trained for each subsamples extracted from samples by using bootstrap sampling method (Hastie et al., 2009b). A prediction is made by averaging the predictions of the trees. This prediction exhibits higher accuracy than that made by a single regression tree. Unfortunately, the trees produced in this approach exhibit a similar tree structure, which is called tree correlation. The higher tree correlation causes a lower prediction performance in the bagging regression trees (Breiman, 2001). Random forest (RF) is an extension of bagging regression trees by incorporating randomized feature selection for reducing the correlation between the trees, as sketched in Fig. 11. RF selects *n* (a predefined number) features among the total *m* features for the split in each node. The RF algorithm tries to find the best split among only the *n* features. The number *n* is set the same for all prediction trees, and it is recommended to be $1/3m$ or $\sqrt{2}\,m$ (Breiman, 2001). The remainder of the algorithm is similar to the CART algorithm.

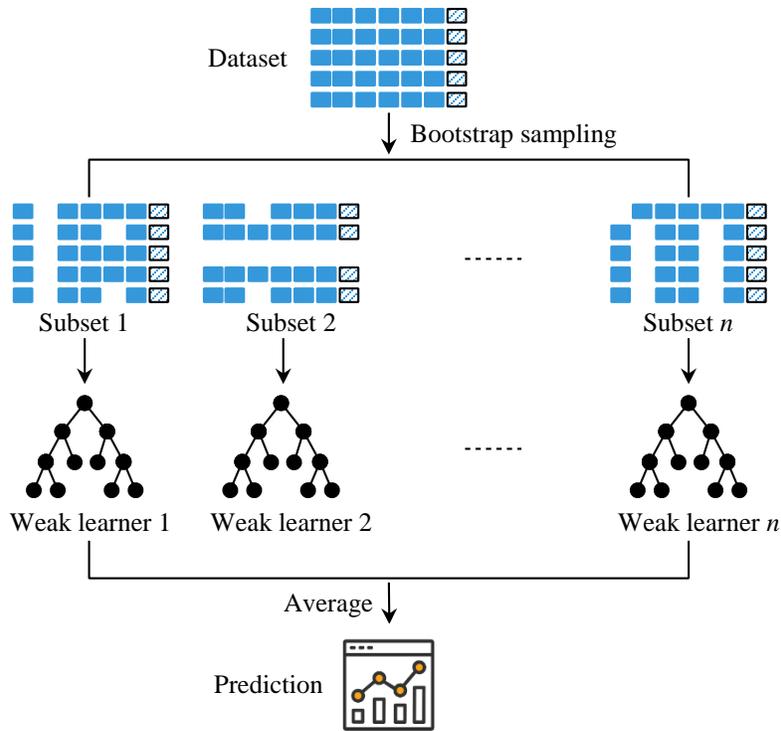

Fig. 11 General procedure of random forest.

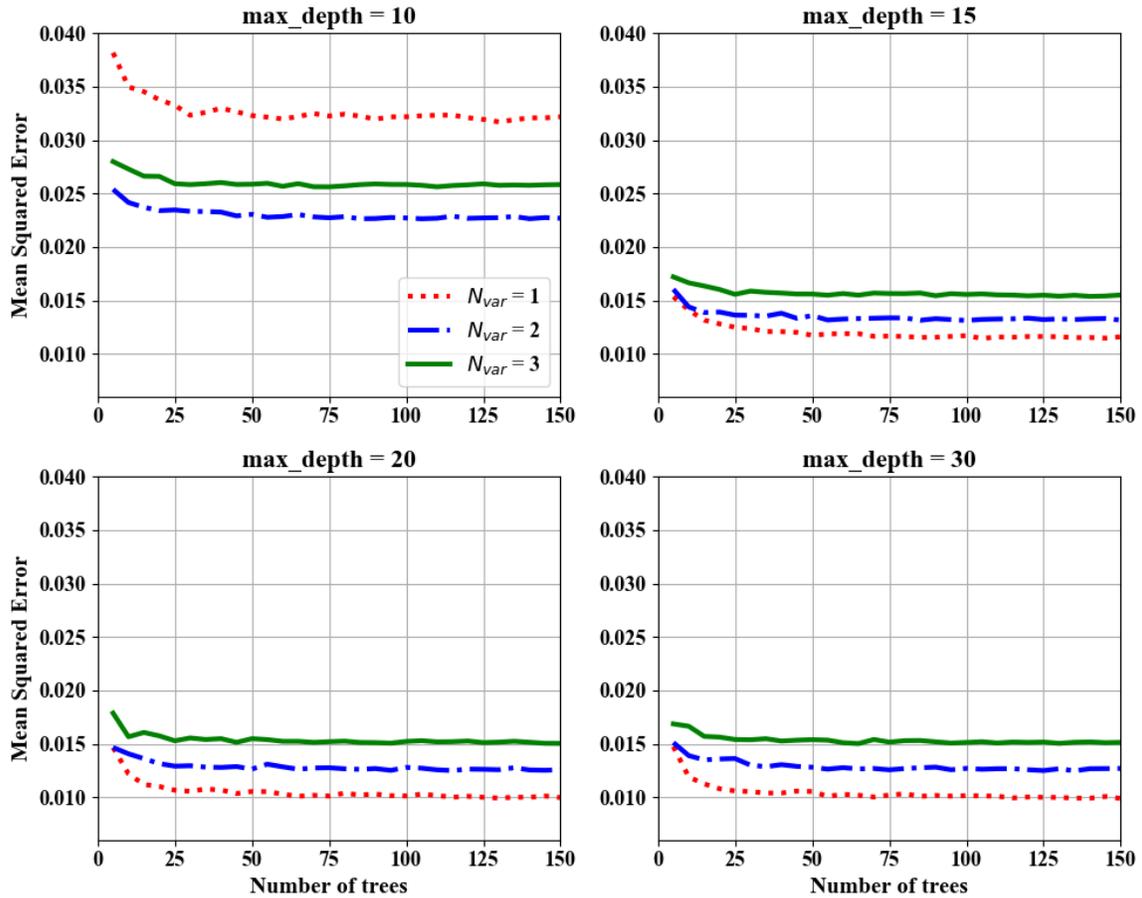

Fig. 12 Variations of 10-fold MSE with hyperparameters of RF for the data set of mean pressure coefficients.

In Fig. 12, variations of 10-fold MSE with number of grown trees, number of selected features, and the maximum depth of trees are presented for the data set of mean pressure coefficients. As can be seen, MSE drops quickly with increasing maximum depth of trees up to 20. As the maximum depth reaches 20 or 30, MSE decreases with reducing the number of selected features. Furthermore, MSE is generally stable as the number of trees exceeds 125. Therefore, the number of grown trees, the number of selected features, and the maximum depth were set as 150, 1, and 20 respectively for the data set of mean pressure coefficients. Similarly, it was found that 150, 2, and 20 are the optimal values for the three parameters respectively for the data set of fluctuating pressure coefficients.

*(b) Boosting: Gradient Boosting Regression Trees (GBRT)*

The main feature of boosting is to add new weak learners to the ensemble sequentially. At each iteration, the ensemble fits a new learner to the difference between the observed response and the aggregated prediction made by all learners grown so far. Boosting regression trees incorporate the strengths of both regression trees, handling various types of predictor variables and accommodating missing data, thus boosting and improving predictive performance by combining many simple learners (Elith et al., 2008). Gradient boosting regression trees (GBRT) is one of the most widely used boosting regression trees and has been recognized as a powerful and successful ML technique in a wide range of practical applications (Natekin and Knoll, 2013; Persson et al., 2017).

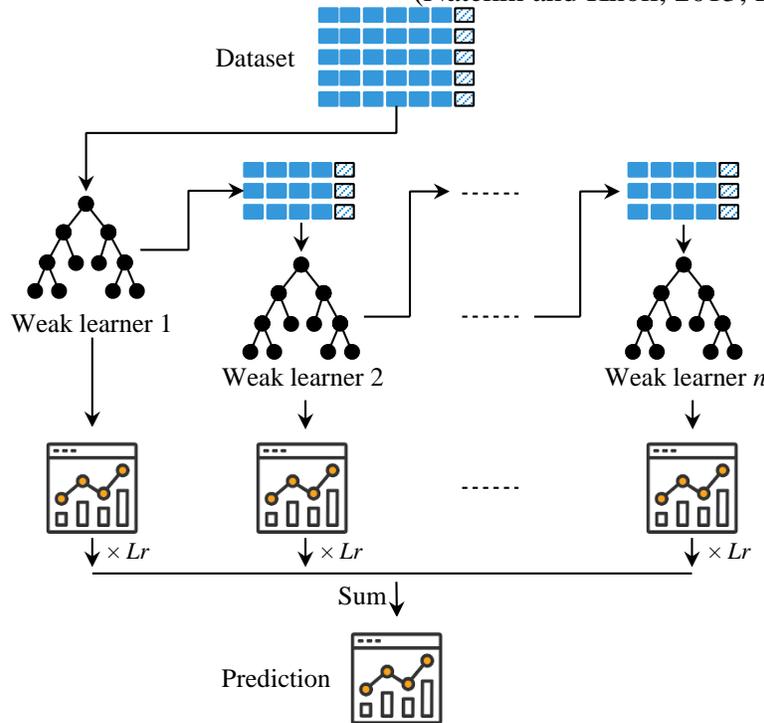

Fig. 13 General procedure of gradient boosting regression trees.

In GBRT, a gradient descent algorithm is used to minimize the squared error loss function. As mentioned above, a new regression tree is fitted to the current residuals at each boosting iteration. After adding enough regression trees to the ensemble, the training error reaches a minimum. To avoid overfitting, the contribution of each regression tree is scaled by a factor, called the learning rate ($Lr$) that is substantially less than 1. In other words, the learning rate determines the contribution of each tree to the final model. The prediction values by the final model are computed as the sum of all trees multiplied by the learning rate (Elith et al., 2008). Another important parameter in GBRT is the maximum tree depth (*Td*) that controls whether interactions are fitted. *Lr* and *Td* determines the number of regression trees (*Nt*) required for the optimal model. Generally, smaller *Lr* increases regression trees for converge. It has been proven by numerous studies that smaller learning rates result in less test error (Natekin and Knoll, 2013; Persson et al., 2017; Zhang and Haghani, 2015), although it increases computational time. For a given *Lr*, fitting more trees with larger *Td* leads to less trees being required for minimum error. Therefore, as *Td* is increased, *Lr* must be decreased if sufficient trees are to be fitted (Elith et al., 2008). In addition, to improve the generation capability of the training model, a subsampling procedure was introduced by Friedman (2002). Specifically, a subsample of the training data, quantified by subsampling fraction *Fs*, is drawn randomly from the full data set for fitting the base learner. Consequently, GBRT optimization involves joint optimization of the above four parameters, i.e., *Lr*, *Td*, *Nt*, and *Fs*. In this study, the influences of all these four parameters on the performance of the GBRT model were evaluated via a 10-fold CV method. The evaluation results with *Fs* =0.3 for the data set of mean pressure coefficients are given in Fig. 14, since *Fs* =0.3 exhibits the best performance compared to 0.1, 0.5, and 0.7 in this study. It was found that the optimal *Lr*, *Td*, and *Nt* are 0.01, 8, and 5000 respectively for the data set of mean pressure coefficients. Similarly, 0.01, 16, 3000, and 0.3 are found to be the optimal values of *Lr*, *Td*, *Nt*, and *Fs* for the data set of fluctuating pressure coefficients.

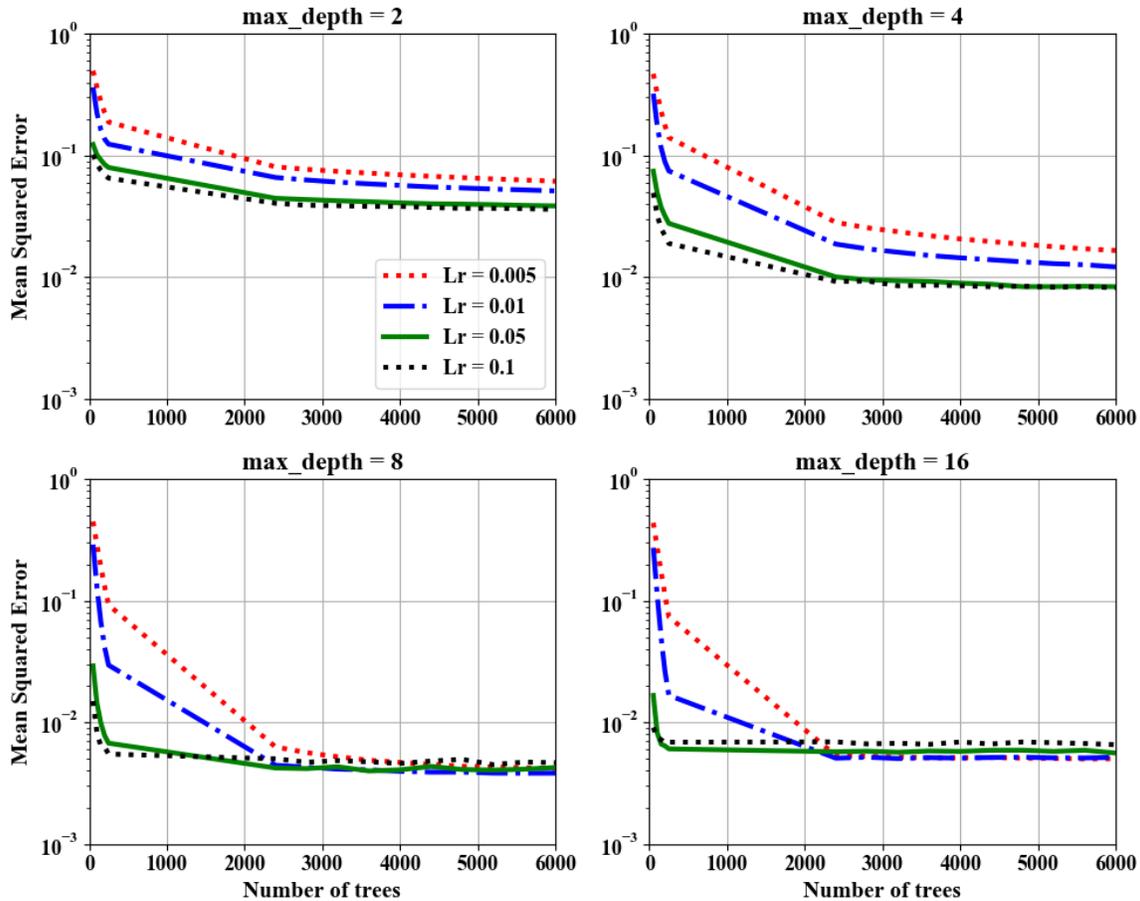

Fig. 14 Variations of 10-fold MSE with hyperparameters of GBRT for the data set of mean pressure coefficients.

## 4. Results and analyses

### 4.1. Performance evaluations of ML models

In the above section, three ML algorithms (DTR, RF and GBRT) have been tested to train models for predicting pressure coefficients around circular cylinders. Meanwhile, their hyperparameters have been optimized to minimize mean squared error (MSE) of the predictions by using 10-fold CV method. So far, the first two stages of the 3-stage evaluation process in Fig. 7 have been accomplished. In this section, three ML models using the three algorithms with their optimal hyperparameters based on the entire training data set are built and tested on the testing data subset.

Comparisons between the predicted mean pressure coefficients and the testing data are given in Fig. 15. Comparisons on a random segment with 100 mean pressure coefficients from the testing data set are shown in Fig. 15(a). In general, all the three ML models are capable of making good predictions. The R-squared ($R^2$) scores of the three models on the test data set in Fig. 15(b) to (d) quantify their performances. Evidently, the two ensemble methods, i.e., RF and GBRT, exhibit a better performance than the single regression tree model (DTR). The GBRT model has the highest score, which implies that this model has the best performance on predicting the mean pressure around circular cylinders. This is further proven by the MSE shown in Fig. 17(a). Likewise, the performances of the three ML models on fluctuating pressure coefficients are compared in Fig. 16 and Fig. 17(b). Similarly, the GBRT model presents the best performance. Therefore, the GBRT models with the optimized hyperparameters are chosen as the ML model for predicting both the mean and fluctuating pressure coefficients.

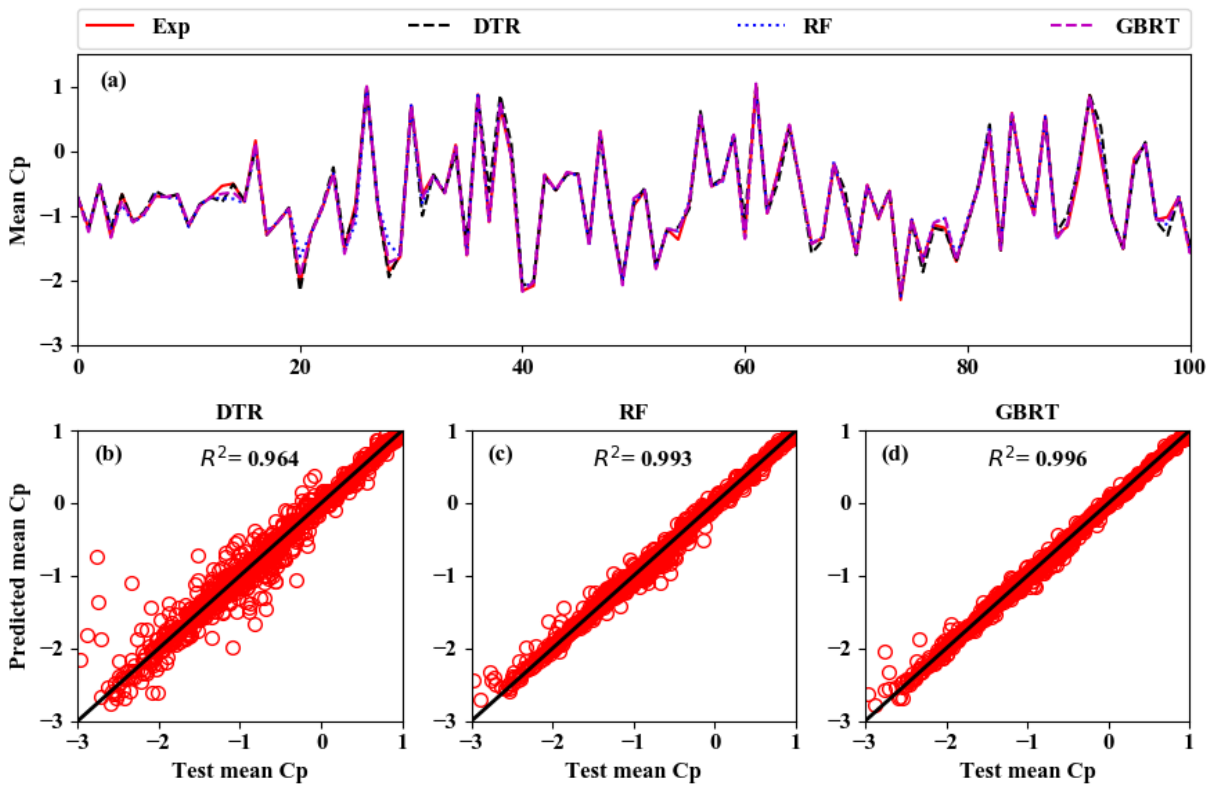

Fig. 15 Comparisons between test mean Cp and mean Cp predicted by ML models.

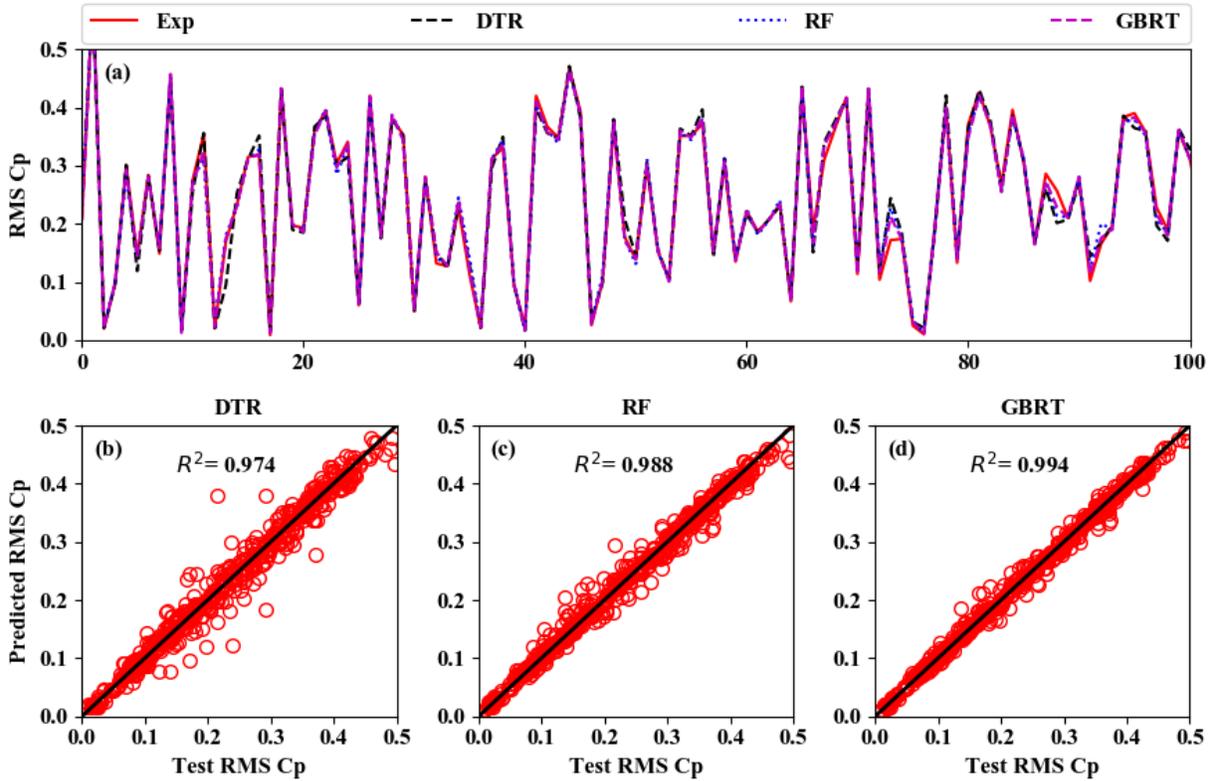

Fig. 16 Comparisons between test fluctuating Cp and fluctuating Cp predicted by ML models.

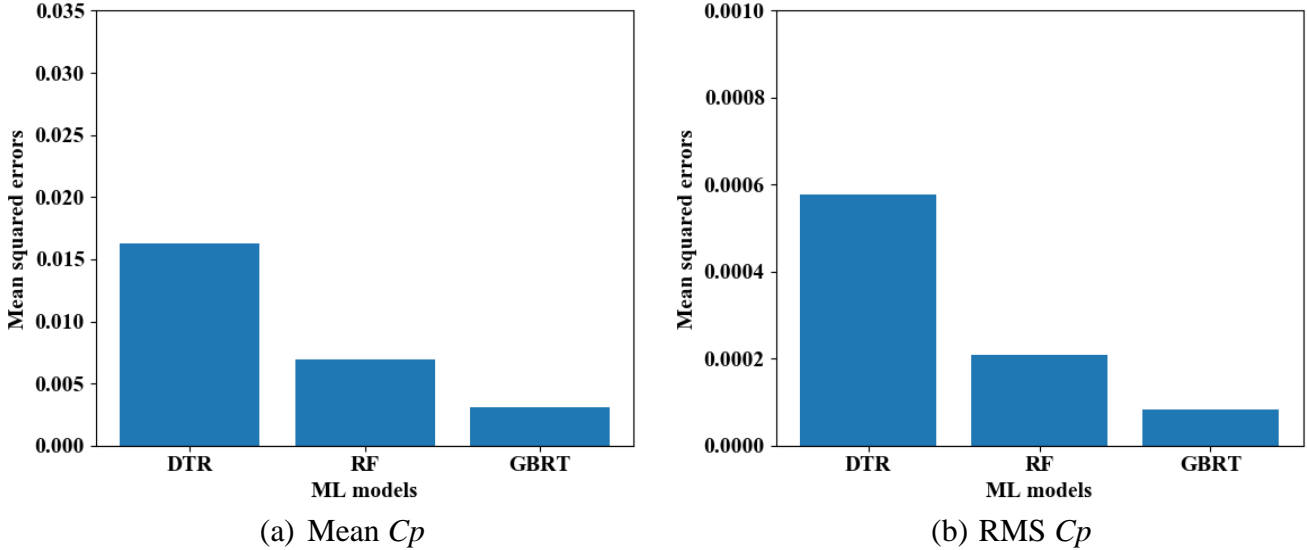

(a) Mean *Cp*  (b) RMS *Cp*

Fig. 17 Comparison of mean square errors of ML models for predicting pressure coefficients.

### 4.2. Final ML models for predicting mean and fluctuating pressure coefficients

The GBRT model with its optimized hyperparameters has been identified as the best model for predicting both mean and fluctuating pressure coefficients in this study. The final prediction models for mean and fluctuating pressure coefficients are built using the GBRT with its optimized hyperparameters based on the whole dataset including both training dataset and testing dataset, i.e., all the blue circle markers in Fig. 6. To further demonstrate the capability of these two final ML models, predictions on the mean and fluctuating pressures around a smooth circular cylinder for their respective 6 sets of *Re* and *Ti* were made and given in Fig. 18 and Fig. 19 respectively. It should be mentioned that the 12 sets of *Re* and *Ti*, i.e. the red triangular markers in Fig. 6, were randomly selected and set aside from the training and testing data set at beginning. Therefore, these data sets are eligible to be used to validate the final ML models.

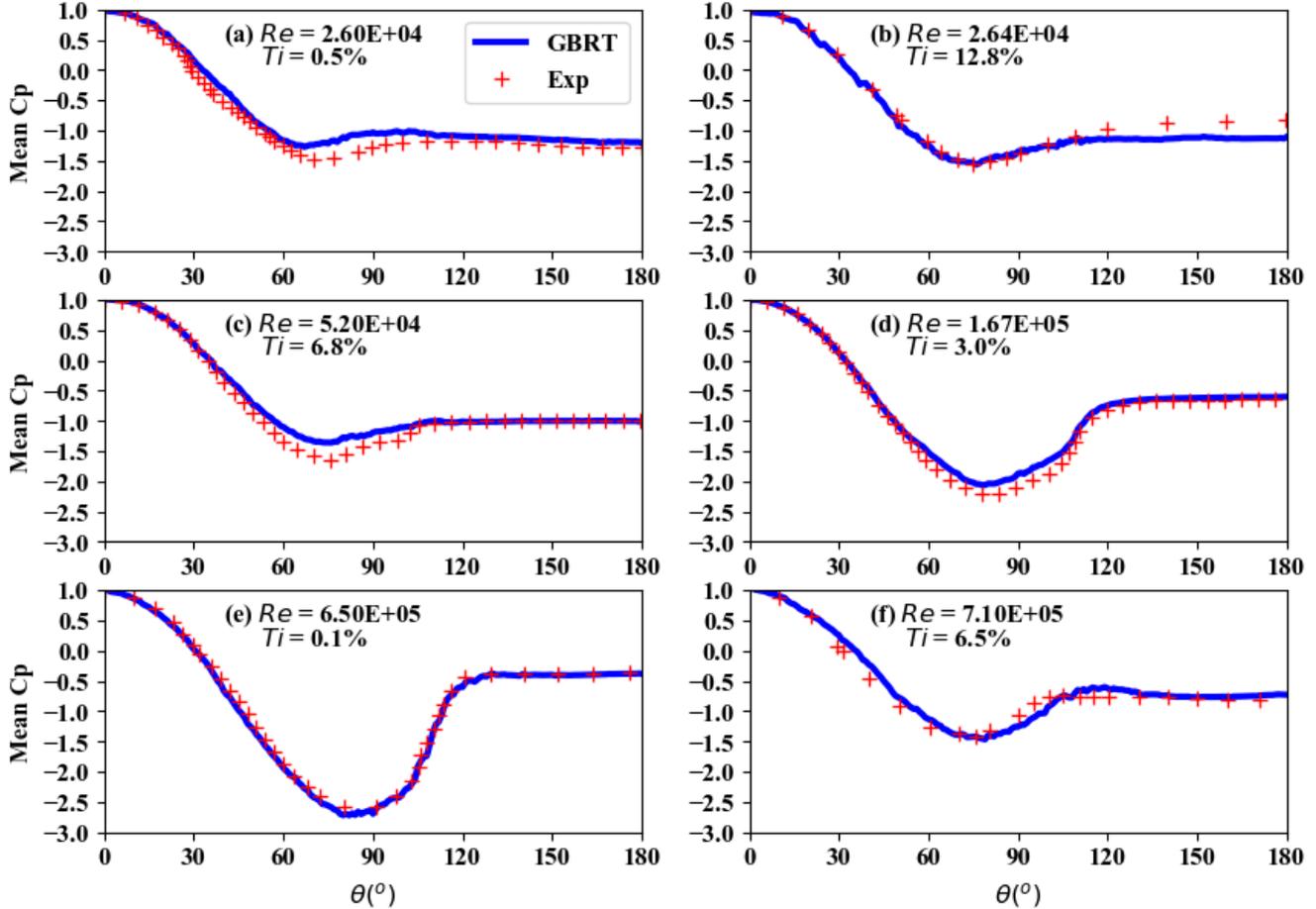

Fig. 18 Prediction of mean pressure coefficients around circular cylinders by using final ML model. Experimental data are from: (a) Igarashi (1986); (b) Kiya et al. (1982); (c) Sadeh and Saharon (1982); (d) Sadeh and Saharon (1982); (e) Sun et al. (1992); (f) Basu (1986).

As can be seen, the GBRT models accurately predict both the mean and fluctuating pressures under their respective 6 sets of *Re* and *Ti*, although minor discrepancies can be found in some cases such as those shown in Fig. 18(a) and Fig. 19(d). Therefore, the two GBRT models are capable of providing accurate predictions of the mean and fluctuating wind pressures around smooth circular cylinder under various combination of *Re* ranging from $10^4$ to $10^6$ and *Ti* ranging from 0% to 15%. It is believed that the GBRT models provide a very efficient and economical alternative to traditional wind tunnel tests and computational fluid dynamic simulations for predicting pressures around smooth circular cylinders in the studied *Re* and *Ti* range.

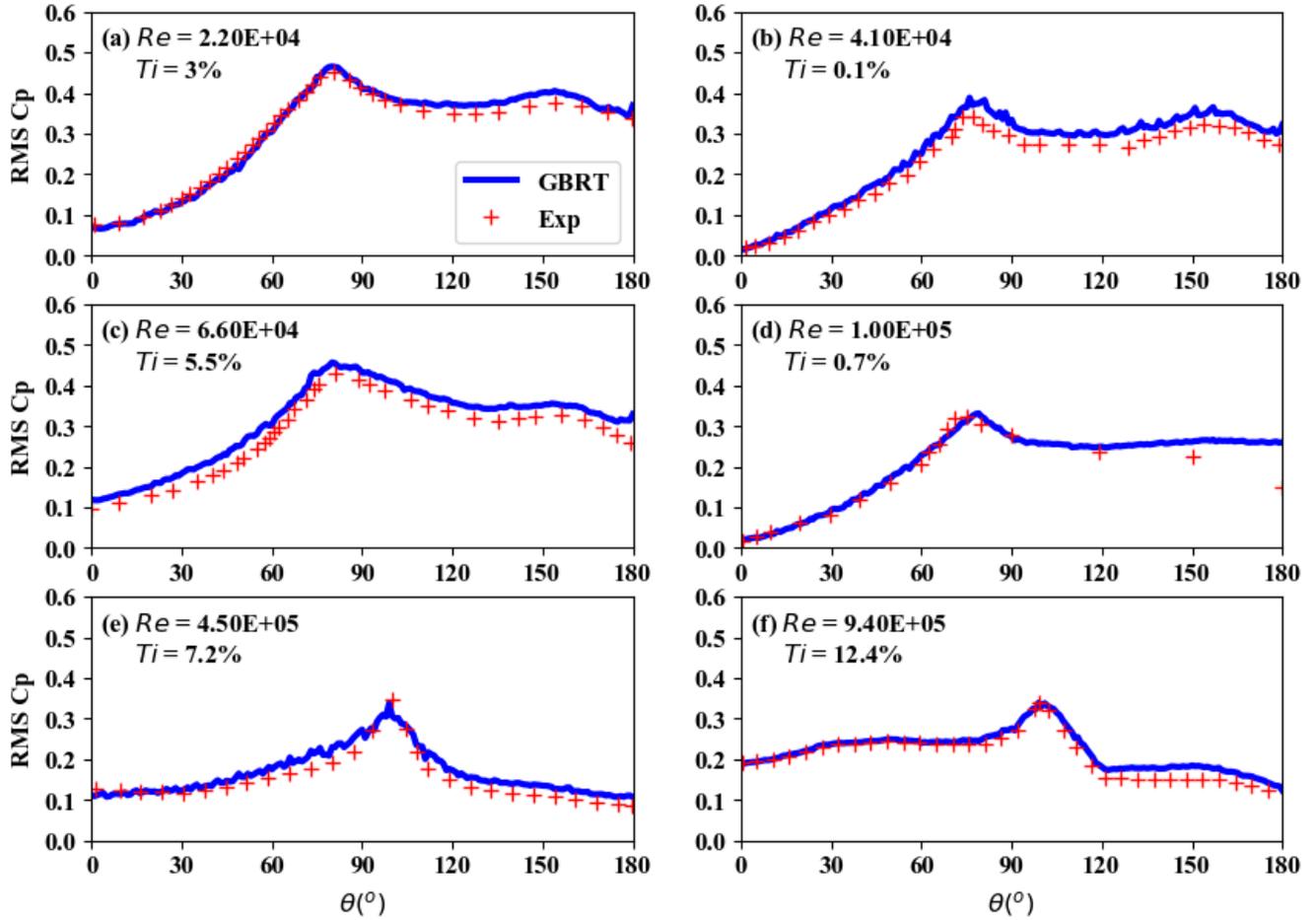

Fig. 19 Prediction of fluctuating pressure coefficients around circular cylinders by using final ML model. Experimental data are from: (a) West and Apelt (1993); (b) Norberg (1986); (c) West and Apelt (1993); (d) Yokuda and Ramaprian (1990); (e) Eaddy (2004); (f) Eaddy (2004).

## 5. Discussions

Machine learning technique has been proven successful in predicting the wind pressures around smooth circular cylinders for $Re$ ranging from $10^4$ to $10^6$ and $Ti$ ranging from 0% to 15% in this study, in spite of minor discrepancies in some cases. It is anticipated that feeding more data to upgrade these models will be able to significantly diminish these discrepancies. Although it has been recognized that surface roughness has a significant effect on the wind pressures around circular cylinders (Güven et al., 1980; Nakamura, 1982), insufficient data addressing this aspect prevents the inclusion of this feature in the ML models. Similarly, the $Re$ ranging from $10^6$ to $10^8$ is of great interest to wind engineers since many practical wind-sensitive structures, such as buildings with circular features, industrial chimneys and cooling towers experience such a high $Re$ flow (Ke et al., 2017; Zhao et al., 2017; Zou et al., 2018). However, the two ML models are unable to cover this $Re$ range due to a limitation of available data. Nevertheless, it is believed that the two ML models can be readily upgraded to address the high $Re$ flow around circular cylinders when relevant data become available in future.

## 6. Concluding remarks

This study has used machine learning (ML) techniques to predict the wind pressures around circular cylinders. The mean and fluctuating pressure data were collected from previous experimental studies presented in literature. Three ML algorithms, including decision tree regressor, random forest, and gradient boosting regression trees (GBRT), have been tested on the data set. It was found the GBRT models exhibit the best performance for predicting both the mean and fluctuating pressure coefficients. Furthermore, the GBRT models are capable of providing accurate predictions on the mean and fluctuating pressure coefficients around smooth circular cylinders under various combinations of Reynolds number (*Re*) ranging from $10^4$ to $10^6$ and flow turbulence intensity (*Ti*) ranging from 0% to 15%. Thus, the GBRT models provide a very efficient and economical alternative to traditional wind tunnel tests and computational fluid dynamic simulations for predicting the wind pressures around smooth circular cylinder in the studied *Re* and *Ti* range.

Although ML technique has been successfully used in various engineering fields and has been proven to have a huge potential in solving practical engineering problems, its application in wind engineering is still in the infant stage. This study has employed ML to predict the mean and fluctuating pressure around a smooth cylinder within specific range of *Re* and *Ti*, and demonstrated the potential applying ML to wind engineering. Therefore, it is worth devoting more efforts to employ ML techniques to address intransigent issues in the wind engineering field, such as high *Re* flow and the effects of surface roughness on the pressure distribution around circular cylinder.